\documentclass[twocolumn]{IETEJR}
\usepackage[square,numbers]{natbib}
\usepackage{hyperref}
\usepackage{graphicx}%
\usepackage{multirow}%
\usepackage{amsmath,amssymb,amsfonts}%
\usepackage{amsthm}%
\usepackage{mathrsfs}%
\usepackage[title]{appendix}%
\usepackage{xcolor}%
\usepackage{textcomp}%
\usepackage{manyfoot}%
\usepackage{booktabs}%
\usepackage{algorithm}%
\usepackage{algorithmicx}%
\usepackage{algpseudocode}%
\usepackage{listings}%
\usepackage[utf8]{inputenc}
\usepackage{float}
\usepackage{csquotes}
\usepackage{subfig}
\usepackage{wrapfig}
\usepackage{array}
\newcolumntype{L}{>{\centering\arraybackslash}m{1cm}}
\usepackage{breqn}

\begin{document}

\title{A Methodology-Oriented Study of Catastrophic Forgetting in Incremental Deep Neural Networks}
\author{ Ashutosh Kumar, Sonali Agarwal, D Jude Hemanth\\
\vspace*{0.05em}\\\small Ashutosh Kumar is with the Indian Institute of Information Technology Allahabad,\\\small Prayagraj, India (corresponding author to provide phone: +91-7309630852;\\\small e-mail: pse2016003@iiita.ac.in). \\
    \small Sonali Agarwal is in the Department of Information Technology at the Indian Institute of Information Technology, Allahabad.\\\small(e-mail: sonali@iiita.ac.in).\\
    \small D Jude Hemanth is in the Department of Electronics and Communication Engineering, KITS, Coimbatore-641114, India.\\\small(e-mail: judehemanth@karunya.edu).\\
    }

\twocolumn[
\begin{@twocolumnfalse}
    \maketitle
    \begin{abstract}
    Human being and different species of animals having the skills to gather, transferring knowledge, processing, fine-tune and generating information throughout their lifetime. The ability of learning throughout their lifespan is referred as continuous learning which is using neurocognition mechanism. Consequently, in real world computational system of incremental learning autonomous agents also needs such continuous learning mechanism which provide retrieval of information and long-term memory consolidation. However, the main challenge in artificial intelligence is that the incremental learning of the autonomous agent when new data confronted. In such scenarios, the main concern is catastrophic forgetting(CF), i.e., while learning the sequentially, neural network underfits the old data when it confronted with new data. To tackle this CF problem many numerous studied have been proposed, however it is very difficult to compare their performance due to dissimilarity in their evaluation mechanism. Here we focus on the comparison of all algorithms which are having similar type of evaluation mechanism. Here we are comparing three types of incremental learning methods: (1) Exemplar based methods, (2) Memory based methods, and (3) Network based method. In this survey paper, methodology oriented study for catastrophic forgetting in incremental deep neural network is addressed. Furthermore, it contains the mathematical overview of impact-full methods which can be help researchers to deal with CF.
    \end{abstract}
\end{@twocolumnfalse}
]
\vspace*{0.05em}
\section{INTRODUCTION}
\label{sec1}
\begin{figure}[htbp]
\centering
\includegraphics[width=0.8\linewidth, height=4cm]{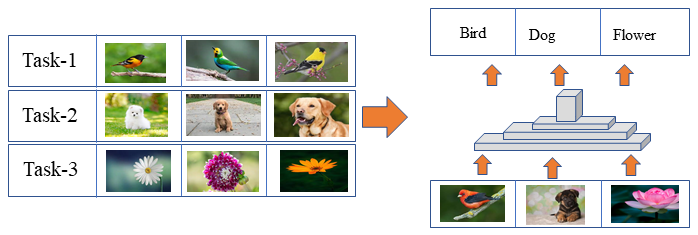}
\caption{Multi-class classifier}
\label{fig11}
\end{figure}

In recent time, deep learning models have surpassed the performance level of humans in classification, object recognition\cite{russakovsky2015imagenet} and Atari games\cite{silver2018general}. However, these task has been done in static scenarios. For continuous learning of any such system we need to develop the artificial intelligent system that can learn new tasks from new confronted data while preserving the old knowledge learned from previous tasks\cite{rebuffi2017icarl}. The learner model should be capable of retaining all its previous knowledge after each time it confronted with new training data. As human confronted with new data each time they tend to learn that information while preserving the old knowledge; similar biological phenomenon is a motivation for artificial intelligence system to integrate the new knowledge with previous learned task. Conventionally, we learn all the task at a single time by aggregating the all data into a single instance- known as multitask \cite{ruder2017overview} as shown in Fig \ref{fig11}.

In Fig. \ref{fig11} we can see that there are three task named as task-1 (birds), task-2 (dog) and task-3 (flower) are aggregated to a single instance and they are learned by a single model. This model classifies the task related to these dataset without any concern. But here is one limitation i.e., the effective use of memory. For learning many task we need more storage and computing power \cite{zhao2021memory}. 

\begin{wrapfigure}{r}{4cm} 
    \includegraphics[width=4cm, height=5cm]{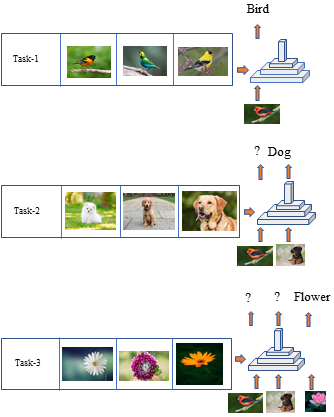}
    \caption{Catastrophic Forgetting Phenomenon}
    \label{fig12}
\end{wrapfigure}

\begin{wrapfigure}{l}{0.2\textwidth} 
    \includegraphics[width=3cm, height=3cm]{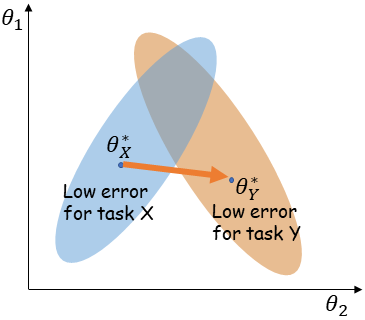}
    \caption{Catastrophic Forgetting}
    \label{fig13}
\end{wrapfigure}

To overcome this problem we can use the incremental learning also referred as online learning in which data continuously comes into chunks and confronted with the learning model\cite{belouadah2021comprehensive,kading2016fine,wu2019large}. In incremental learning scenario we do not need to store data as in conventional multitask approach\cite{rebuffi2017icarl,kirkpatrick2017overcoming}.
However, incremental deep neural network models tends to forget the old learned knowledge when it confronted with the new data - the phenomenon is known as catastrophic forgetting\cite{kirkpatrick2017overcoming,goodfellow2013empirical} as shown in Fig. \ref{fig12}. We can see catastrophic forgetting in Fig. \ref{fig12} where task-1(bird dataset) learned through the model it acquires the knowledge of task-1 and classifies its data. But when same model learned with task-2(dog dataset), it forget the previous task and acquires new weights to model and only classifies the instances of dog dataset. Similarly, when model confronted with task-3(flower dataset) it tend to forget the previous tasks. This phenomenon is known as \enquote*{catastrophic forgetting}. The mathematical representation of catastrophic forgetting is described in Fig. \ref{fig13}.
The mathematical representation of catastrophic forgetting is described below:\\
Here, for each classification task, $D = (Z,Z')$ represents the dataset which contains inputs to $Z = (z_1,z_2,z_3,....,z_n)$ and respective labels $Z'= (z'_1,z'_2,z'_3,....,z'_n)$. The classification task can be described as learning of posterior probability as $p(\theta \mid D)$, where $D$ is a dataset and $\theta$ is DNN parameter. It can be also represented as logarithmic of Bayes' theorem as indicated in Eq. \ref{eq11}:
\begin{equation}
 \log p\left ( \theta \mid D \right ) = \log p\left ( D \mid \theta \right ) + \log p\left ( \theta  \right ) -\log p\left ( D \right )
 \label{eq11}
\end{equation}
where $log\ p(\theta \mid D)$ is posterior, $log\ p(D \mid \theta)$ is likelihood and $log \ p(\theta)$ is prior.

During training of a DNN, the loss is minimized for maximum likelihood estimation and the prior is neglected as represented in equation \ref{eq12}, where to classify the task cross-entropy loss is used.
\begin{equation}
    L(\theta) = -log \ p(D \mid \theta) = - log \ p(Z'\mid Z,\theta)
    \label{eq12}
\end{equation}
Now, suppose the dataset $D$ is divided into two disjoint sets of tasks $D_X$ and $D_Y$ for two different tasks $X$ and $Y$ respectively. These two different sets can be considered as two separate classification tasks: $X$ and $Y$. Let the same DNN is utilized to learn the task $X$ followed by $Y$ without reconsidering the task $X$ samples. For task $X$, the maximum likelihood solution is $\theta^*_X$ by minimizing the loss as shown in equation \ref{eq13}:
\begin{equation}
    \theta^*_X = arg \ \underset{\theta}{max}  \ \  L_X(\theta)
    \label{eq13}
\end{equation}
Similarly, the second task $Y$ is learned in sequential fashion for respective dataset as shown in equation \ref{eq14}:
\begin{equation}
    \theta^*_Y = arg \  \underset{\theta}{max}  \ \   L_Y(\theta;\ \theta^*_X)
    \label{eq14}
\end{equation}
where inclusion of $\theta^*_X$ explicitly indicates that the DNN is initialized with resulting parameter of recent task. 

The above discussed idea forgets the task $A$ while learning the task $Y$. Fig. \ref{fig13} depicts the overall scenario, where it is assumed that the model has only two parameters $\theta_1$ and $\theta_2$ for training.


For specific task $X$ in a learning model, $\theta^*_X$ corresponds to a low error: $L_X(\theta)$ (the gray shaded region). To remember task $X$ we want to stay in that region. Similarly, in blue shaded region represents the model for task $Y$ where loss is low enough with respect to task $Y$. But, when learning of task $Y$ is performed the the gradient descent will move $\theta$ from the initial $\theta^*_X$ toward the optimal solution $\theta^*_Y$.

 

As discussed by the authors that the above solution can be achieved by \enquote*{Elastic Weight Consolidation(EWC)} approach \cite{kirkpatrick2017overcoming}. Given the above approximation, we minimize the loss function for EWC as given in equation\ref{eq15}
\begin{equation}
    \mathscr{L}(\theta) =  \mathscr{L}_{Y}(\theta) + \sum_{i}\frac{\lambda }{2}F_{i}\left ( \theta _{i}-\theta _{X,i}^{*} \right )^{2}
    \label{eq15}
\end{equation}
where loss for task $Y$ is $\mathscr{L}_{Y}(\theta)$, the importance of previous task in compared with new one is described by $\lambda$ and label of each parameter is defined by $i$.

Our survey for comparing the state of the art approach for incremental learning to handle catastrophic forgetting based on network architecture as well as the efficiency of the algorithms. Our contribution is as follows:
\begin{itemize}
    \item Our survey compares the existing approaches based on the their Complexity, Accuracy, Plasticity, Memory, Timeliness and Scalability.
    \item The comparison of existing approach is also based on the behaviour of the algorithm like Fixed-Representation based, Fine-Tuning based and Model-Growth based. 
\end{itemize}

Rest of the paper is organised as follows: In Section \ref{sec2}, it contains a brief discussion of existing incremental deep learning frameworks, and Section \ref{sec3} discusses the learning scenarios. Methods for incremental learning are discussed in Section \ref{sec4}, whereas mathematical overview of methods are described in Section \ref{sec5}, Section \ref{sec6} described the conclusion and future work is discussed in section \ref{sec7}.

\section{Continual Learning Frameworks}
\label{sec2}
In the discipline of machine learning, as well as in other domains that are linked to it, research has been done in the previous on the concepts that are illustrated in the continuous learning criteria of knowledge transfer, sharing, and adaptation. In this section, we will provide a concise explanation of each of them, focusing on the primary distinctions between them related to continual learning setting. Different continual learning frameworks are described as below:
\begin{enumerate}
\item \textbf{Standard supervised learning:} Supervised learning is a method of developing artificial intelligence (AI) in which a computer programme is trained on labelled input data for a certain output. As shown in figure Fig. \ref{fig21:subim1} where task $T_1$ is having dataset $(X,Y) \in D_1$ means that all the datasets are provided at a single instance and model learns the knowledge in same way.

\begin{figure}[htbp]
\centering
\subfloat[Standard Supervised Learning]{\includegraphics[width=0.3\linewidth, height=4cm]{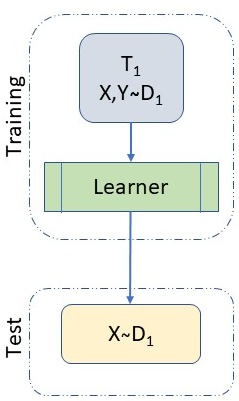} \label{fig21:subim1}} \quad
\subfloat[Multi Task Learning]{\includegraphics[width=0.3\linewidth, height=4cm]{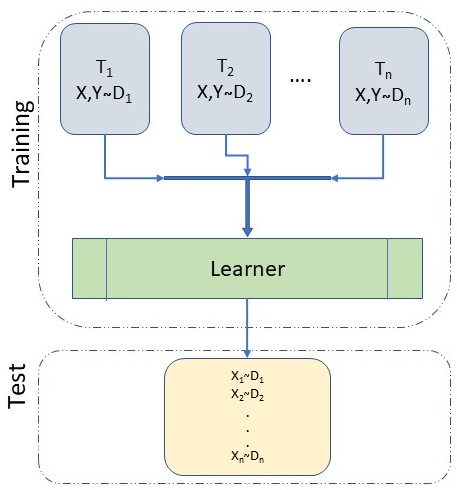}\label{fig21:subim2}}
\caption{Conventional deep learning frameworks}
\label{fig21}
\end{figure}

\item \textbf{Multi task learning:} Multi task learning optimises many loss functions at once to learn various tasks concurrently. Instead of training separate models for each task, we train one model to do everything. In this process, the model leverages all available data across tasks to learn generic data representations. As shown in figure Fig. \ref{fig21:subim2} where task $T \in \{T_1, T_2, ...,T_k \}$ are associated with one task and and model learns the same.

\item \textbf{Domain adaption learning:} Domain adaptation is a technique for improving the performance of a model on a target domain that has insufficient annotated data by making use of the knowledge learned by the model from another related domain that has adequate labelled data. This is accomplished by applying the information gained by the model from the first domain to the second domain as shown in figure Fig. \ref{fig22:subim1}.
\begin{figure}[htbp]
\centering
\subfloat[Domain Adaption Learning]{\includegraphics[width=0.35\linewidth, height=4cm]{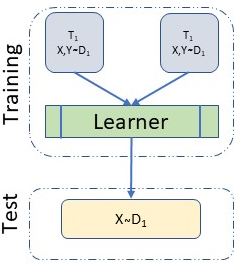} \label{fig22:subim1}} \quad
\subfloat[Transfer Learning]{\includegraphics[width=0.45\linewidth, height=4cm]{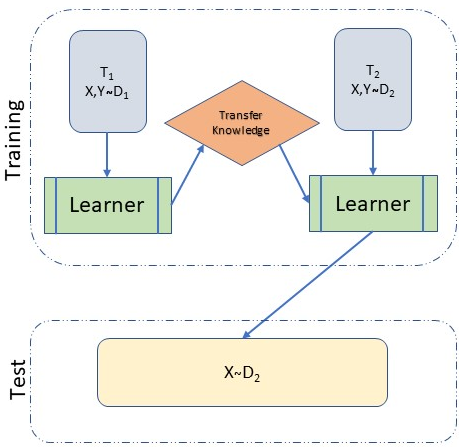}\label{fig22:subim2}}
\caption{Knowledge Sharing learning frameworks}
\label{fig22}
\end{figure}

\item \textbf{Transfer learning:} Transfer learning is a technique of machine learning in which a model that was generated for one task is utilised as the starting point for a model that is being produced for a different task. The concept of transfer learning refers to the process of transcending the isolated learning paradigm and making use of information gained for one activity to tackle problems associated with other tasks as shown in figure Fig. \ref{fig22:subim2}.


\item \textbf{Continuous Incremental Learning:} The idea behind continuous learning, also referred to as incremental learning or life-long learning, is to learn a model for many different tasks sequentially without forgetting the information acquired from the tasks that came before them, even when the data for the older tasks is no longer available when training the new ones. 
\begin{figure}[htbp]
\centering
\includegraphics[width=0.4\linewidth, height=4cm]{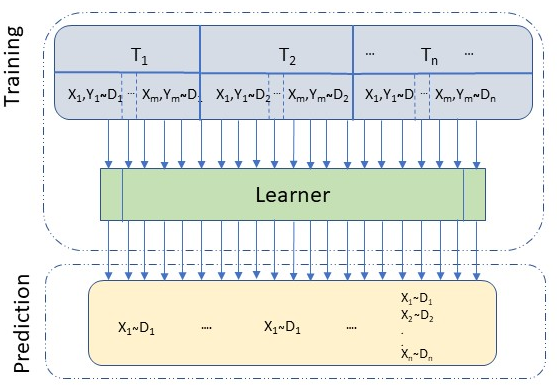}
\caption{Continuous Incremental Learning}
\label{fig25}
\end{figure}

As shown in figure Fig. \ref{fig25} where we can see different task are encountered sequentially to model and no previous task data is available while learning new tasks however model preserves old task knowledge and acquires new task knowledge with some forgetting.

\item \textbf{Online learning:}Online learning is a blend of many ML approaches in which data comes sequentially and the learner seeks to learn and improve the best predictor for future data at each step. 

\begin{figure}[htbp]
\centering
\subfloat[Online Learning]{\includegraphics[width=0.35\linewidth, height=4cm]{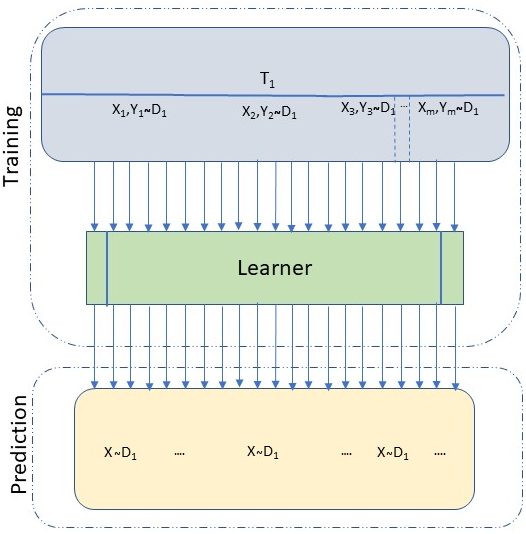} \label{fig23:subim1}} \quad
\subfloat[Meta Learning]{\includegraphics[width=0.45\linewidth, height=4cm]{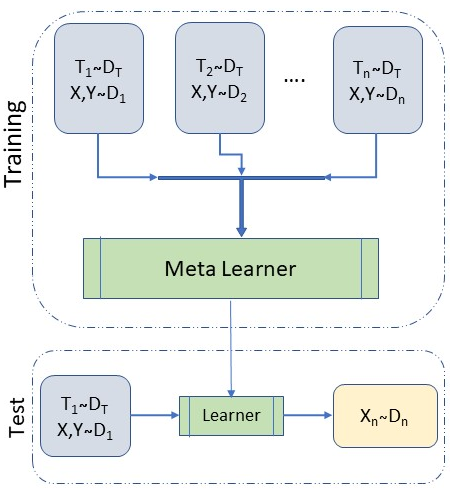}\label{fig23:subim2}}
\caption{Online and Meta Learning}
\label{fig23}
\end{figure}

As a result, for large-scale learning tasks and diverse applications where data is not only vast in quantity but also comes at high velocity as shown in figure Fig. \ref{fig23:subim1}, online learning is significantly more efficient and scalable.

\item \textbf{Meta learning:} In the context of machine learning, the term meta-learning most often refers to ML algorithms that learn by analysing the results of other ML algorithms. 
It is a self-learning algorithm that improves the efficiency of pre-existing algorithms or learns a new algorithm by analysing metadata as shown in figure Fig. \ref{fig23:subim2} and recognising how learning might become flexible in identifying the challenges.
\end{enumerate}

\section{Learning Scenarios}
\label{sec3}
In recent years, various approaches have been proposed to handle catastrophic forgetting in incremental learning scenario. However, due to wide range of datasets and variety of experiments and protocols to overcome catastrophic forgetting many of them claim \enquote*{state-of-the-art} performance \cite{rebuffi2017icarl,kirkpatrick2017overcoming,masse2018alleviating,nguyen2017variational,kemker2017fearnet,wu2018incremental}.
This paper refers to three distinct scenarios in which we have been comparing different methods of incremental learning to alleviate the catastrophic forgetting \cite{van2019three}. 
In incremental learning, we focus on a neural network model which learns series of tasks sequentially. At the time of training, only current task's data is available. Since every time of training, new data in confronted with the model it tend to forget the previous task and learns new task. This phenomenon is known as catastrophic forgetting. In recent years, this problem has been actively studied by various research scholars\cite{kirkpatrick2017overcoming,rebuffi2017icarl,wu2019large}. However, due to various differences in experimental setting used for evaluation, it is very difficult to compare the performance of methods. To tackle this problem Gido and Andreas \cite{van2019three} have proposed a \enquote*{Three scenarios for continual learning} as shown in Table \ref{tab:31}. The three scenarios for continual learning is based on task-ID. As described in Table \ref{tab:31}, it illustrate that three continual learning scenarios are: 

\begin{table}[htbp]
\centering
\caption{Incremental learning scenarios}
\setlength{\tabcolsep}{5.0pt}
\renewcommand{\arraystretch}{1.5}
\label{tab:31}
\begin{tabular}{ll}
\hline
\textbf{Scenario}                          & \textbf{Test time requirement} \\ \hline
Task based IL*  & task-ID is required            \\
Domain based IL & task-ID not required           \\
Class based IL  & task-ID not required             \\ \hline
\end{tabular}
\end{table}
*IL-Incremental Learning

\begin{figure}[htbp]
\centering
\includegraphics[width=0.8\linewidth, height=4cm]{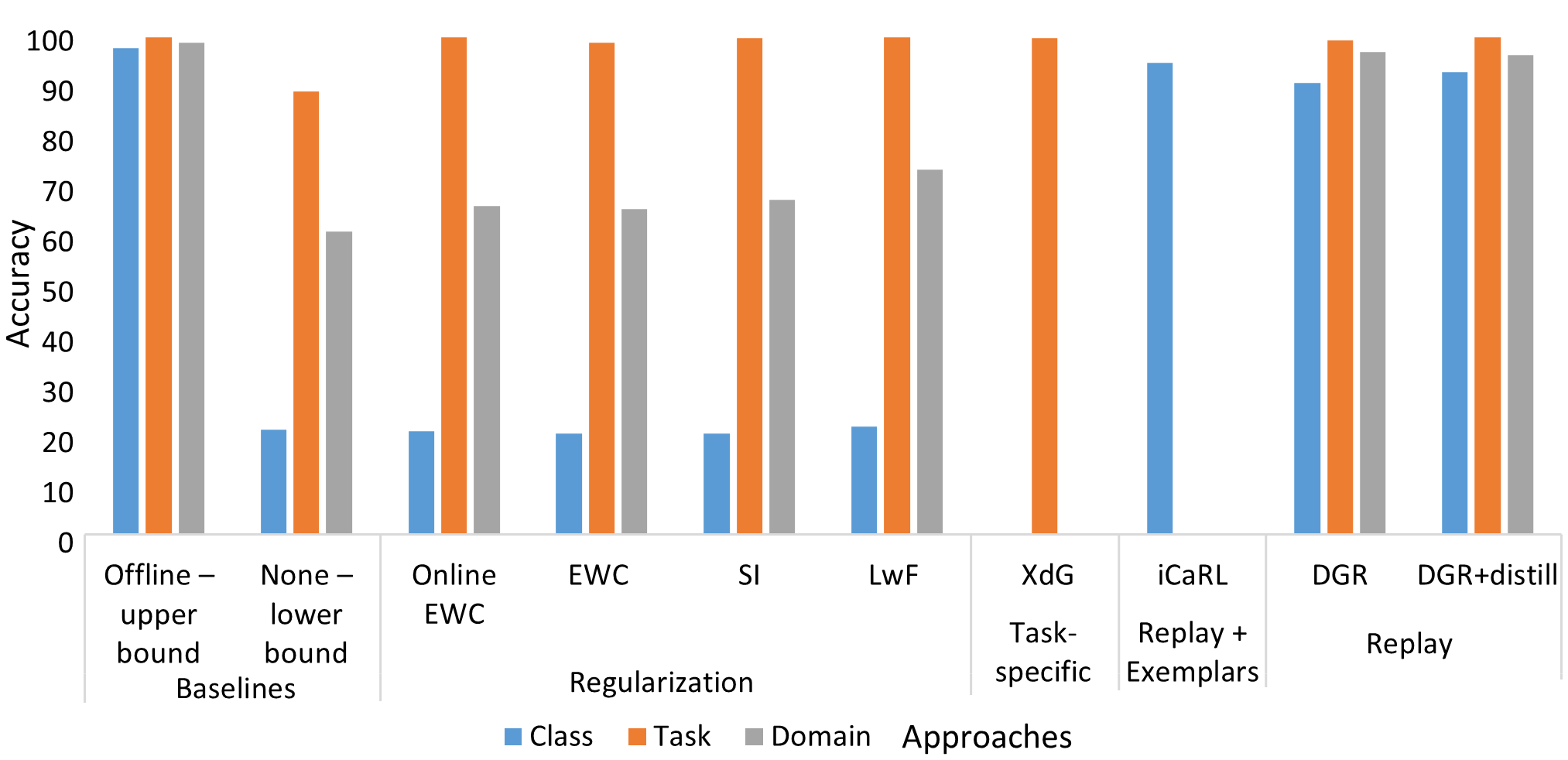}
\caption{Average Validation Accuracy on split MNIST task dataset}
\label{fig31}
\end{figure}

\begin{figure}[htbp]
\centering
\includegraphics[width=0.8\linewidth, height=4cm]{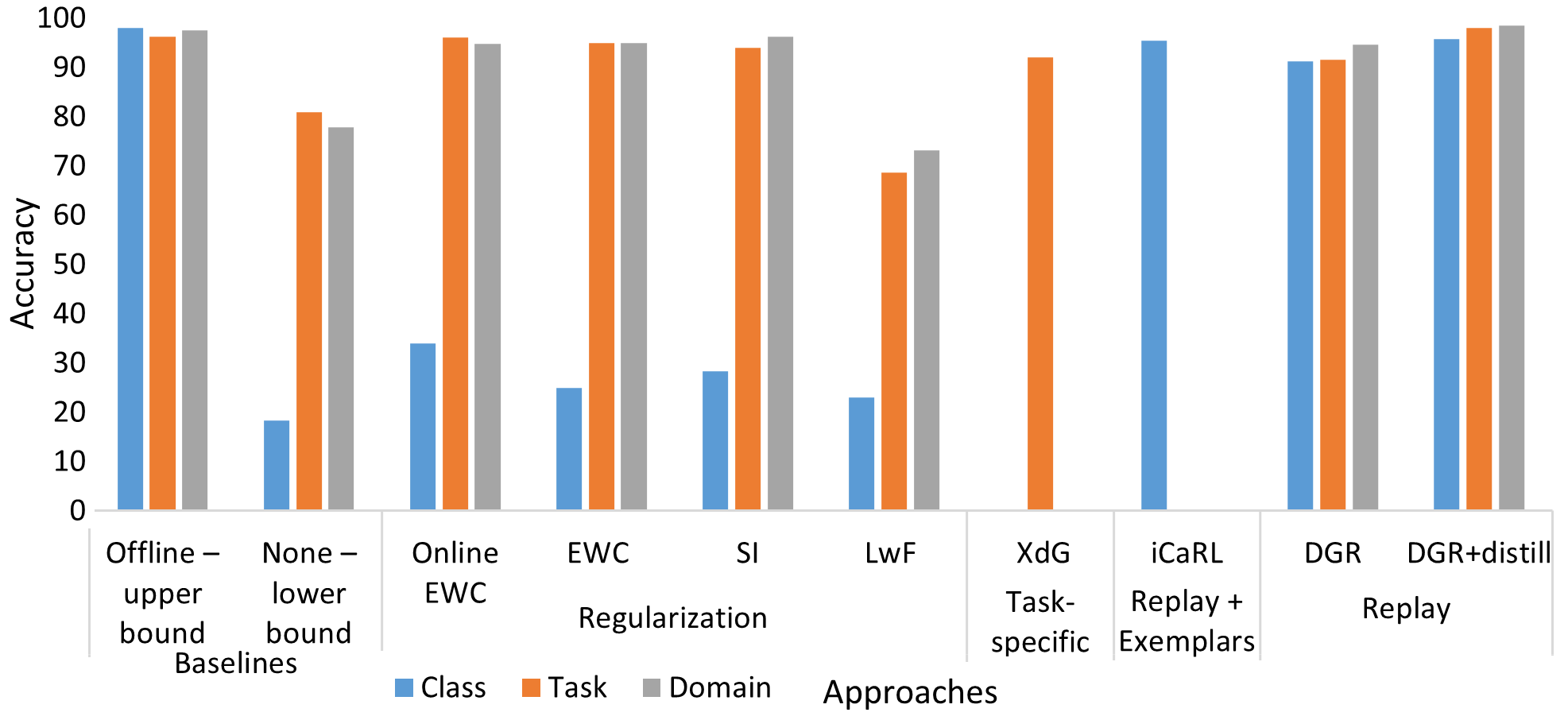}
\caption{Average Validation Accuracy on p-MNIST task dataset}
\label{fig32}
\end{figure}

\begin{enumerate}
\item \textbf{Task based incremental learning:} This scenario solves the task and at test time task-ID is to be provided. In this scenario model always needs the information of which task to be performed. In this scenario \enquote{multi-headed} output layer is needed and each task refer to its respective output units and remaining network potentially shared the weights among tasks. 
\item \textbf{Domain based incremental learning:} This scenario solves the task and at test time task-ID is not required. In this scenario models solve the task as they confronted, model does not required the task information. Here, input distribution is changing every time while task structure is always same. 
\item \textbf{Class based incremental learning:} This scenario solves the task and at test time task-ID is infer to which task they are presented with. In this scenario model incrementally learns new classes which  includes common real world problems.
\end{enumerate}
To illustrate above learning scenarios Ven et.al. \cite{van2019three} have used the MNIST-digits dataset. The first task protocol is used as \enquote*{split MNIST} \cite{zenke2017continual}. This task has been performed under all three incremental learning scenarios as shown in Table \ref{tab:32} and in figure \ref{fig31}. Similarly, for second illustration authors used \enquote*{permuted MNIST} \cite{goodfellow2013empirical}. These task has been performed for same methods as shown in Table \ref{tab:33} and in figure \ref{fig32}.

\begin{table}[ht]
\tiny
\caption{Average Validation Accuracy on split MNIST task dataset}\label{tab:32}
\renewcommand{\arraystretch}{1.5}
\begin{tabular*}{0.5\textwidth}{LLlll}
\toprule%
Approach & Method & Class & Tasks & Domain \\
\midrule
\multirow{2}{4em}{Baseline} & Offline–upper bound & 96.80 ($\pm$ 0.60) & 98.99 ($\pm$0.10) & 97.95 ($\pm$0.50)\\
& None–lower bound & 20.80 ($\pm$0.30) & 88.20 ($\pm$0.64) & 60.35 ($\pm$1.05)\\
\midrule
\multirow{4}{4em}{Regularization} & Online EWC & 20.56 ($\pm$0.03) & 98.95 ($\pm$0.02) & 65.35 ($\pm$0.95)\\
& EWC  & 19.99 ($\pm$0.08) & 97.95 ($\pm$0.38) & 64.75 ($\pm$0.90)\\
& SI & 20.01 ($\pm$0.24) & 98.90 ($\pm$0.75) & 66.56 ($\pm$1.90) \\
& LwF & 21.45 ($\pm$1.25)  & 98.95 ($\pm$0.25) & 72.60 ($\pm$1.05) \\
\midrule
 Task-specific & XdG & - & 98.90 ($\pm$0.95) & - \\
\midrule
 Replay with Exemplars & iCaRL & 93.97 ($\pm$0.58) & - & - \\
 \midrule
\multirow{2}{4em}{Replay} & DGR & 89.85 ($\pm$0.21) & 98.40 ($\pm$0.40) & 96.05 ($\pm$0.55)\\
& DGR+distill &  92.05 ($\pm$0.22) & 99.05 ($\pm$0.50) & 95.45 ($\pm$0.70)\\
\midrule
\end{tabular*}
\end{table}

In incremental learning to alleviate catastrophic forgetting an approach is to assign randomly which nodes participate. In this approach Task-ID is required as used in  \emph{XdG (Context dependent Gating)}\cite{masse2018alleviating}. As in second learning scenario when Task-ID is not required, is executed by regularizing the network while training new task. This approach is used in \emph{SI (Synaptic Intelligence)} \cite{zenke2017continual}, \emph{EWC (Elastic Weight Consolidation)} \cite{kirkpatrick2017overcoming} and \emph{Online EWC} \cite{schwarz2018progress}. Another strategy to overcome catastrophic forgetting is to use the \enquote{Exemplar Data} also referred as \enquote{pseudo data}. In this strategy the representative data of previous task is used with new task data in training \cite{hou2018lifelong}. This strategy is widely used in \emph{LwF (Learning without Forgetting)} \cite{li2017learning} and distilling the knowledge from a larger network to smaller one \cite{hinton2015distilling}. Another approach in which model generates the input data which are available to new training task \cite{cong2020gan}. In these approaches a separate model along with main is incrementally trained on each task to generate the samples form input data distributions, is known as generative model. Many authors used this approach in their methods such as \emph{DGR (Deep Generative Replay)} \cite{shin2017continual} and hybrid model \emph{DGR+distill} \cite{wu2018incremental,venkatesan2017strategy}.

\begin{table}[ht]
\tiny
\caption{Average Validation Accuracy on permuted MNIST(p-MNIST) task dataset}\label{tab:33}
\renewcommand{\arraystretch}{1.5}
\begin{tabular*}{0.5\textwidth}{LLccc}
\toprule%
Approach & Method & Class & Tasks & Domain \\
\midrule
\multirow{2}{4em}{Baseline} & Offline–upper bound & 98.05 ($\pm$0.50) & 96.28 ($\pm$0.25) &
  97.65 ($\pm$0.66)\\
& None–lower bound & 18.37 ($\pm$0.54) & 80.95 ($\pm$1.20)  & 77.91 ($\pm$0.10)\\
\midrule
\multirow{4}{4em}{Regularization} & Online EWC & 34.02 ($\pm$0.20) & 96.12 ($\pm$0.21) & 94.87 ($\pm$0.26)\\
& EWC  & 24.94 ($\pm$1.20)     & 95.06 ($\pm$0.02)    & 95.05 ($\pm$1.33) \\
& SI & 28.42 ($\pm$0.74)     & 93.95 ($\pm$1.18)    & 96.33 ($\pm$0.02) \\
& LwF & 23.05 ($\pm$0.41)     & 68.75 ($\pm$1.05)    & 73.20 ($\pm$0.32) \\
\midrule
 Task-specific & XdG & - & 92.04 ($\pm$0.61) & - \\
\midrule
 Replay with Exemplars & iCaRL & 95.55 ($\pm$0.27) & - & - \\
 \midrule
\multirow{2}{4em}{Replay} & DGR & 91.28 ($\pm$0.07) & 91.56 ($\pm$0.78) & 94.69 ($\pm$0.50)\\
& DGR+distill & 95.78 ($\pm$1.54)     & 98.05 ($\pm$0.70)    & 98.55 ($\pm$0.43)\\
\midrule
\end{tabular*}
\end{table}

If we are having enough memory to store the data from previous task, another approach to overcome catastrophic forgetting is use to store data during execution. These methods uses \enquote{exemplars} for training with new task. In a recent method this approach is used is known as \emph{iCaRL}\cite{rebuffi2017icarl} and in \emph{Large Scale Incremental Learning} \cite{wu2019large}.

The observation from above scenarios for split MNIST task can be seen in Table \ref{tab:32}. We have observed that in Task based incremental learning scenario all the methods performed satisfactory, however EWC, Online EWC and SI (regularization based methods) and LwF did not performed well in Domain based incremental learning scenario and these methods for Class based incremental learning completely failed. One observation to note that replay methods such as DGR and DGR+distill performs well in all three scenarios and iCaRL for Class based incremental learning.

Similarly, for permuted MNIST task all the methods performs well for Task based incremental learning scenario and Domain based incremental learning scenario except for LwF as shown in Table \ref{tab:33}. In the Class based incremental learning scenario replay based methods such as DGR and DGR+distill performs well. On the basis of above observation we can say that replay based methods perform well in all incremental learning scenario.

\section{Incremental Learning Approaches}
\label{sec4}
In section \ref{sec3}, we have described scenarios for overcoming the catastrophic forgetting in incremental learning. Now, in this section \ref{sec4} we will discuss about the different approaches which are being used in recent years to tackle with the catastrophic forgetting problem. We are classifying the approaches based on their usage of stored knowledge and task information in incremental learning. Different approaches and their sub-categorization is given in Table \ref{tab:41}. Here, we have consider approaches such as: architecture based method, replay based methods and regularization based methods with their categorization as shown in Table \ref{tab:41}. Our work is similar to some extent as introduced in \cite{aljundi2019online,farquhar2018towards,van2019three}. However we have introduced latest work in this field and it covers more general overview of existing works.

\begin{table}[ht]
\tiny
\caption{Method Based Classification}\label{tab:41}
\setlength{\tabcolsep}{0.5pt}
\begin{tabular*}{0.5\textwidth}{lLlLLLL}
\toprule%
\multicolumn{3}{@{}c@{}}{Replay} &  \multicolumn{2}{@{}c@{}}{Regularization} &  \multicolumn{2}{@{}c@{}}{Architectural} \\
\cmidrule{1-3}\cmidrule{4-5}\cmidrule{6-7}
Rehersal & Pseudo Rehersal & Constrained & Prior Focused & Data Focused & Fixed Network & Dynamic Network \\
\midrule
TEM \cite{chaudhry2019continual} & LGM \cite{ramapuram2020lifelong} & GSS \cite{aljundi2019online}  & EWC \cite{kirkpatrick2017overcoming} &
LwF \cite{li2017learning} & PathNet \cite{fernando2017pathnet} &
DAN \cite{rosenfeld2018incremental} \\ 

SER \cite{isele2018selective} & PR \cite{atkinson1802pseudo} & GEM \cite{lopez2017gradient} & R-EWC \cite{liu2018rotate}  & DMC \cite{zhang2020class} & PackNet \cite{mallya2018packnet} & Expert Gate \cite{aljundi2017expert} \\

iCarl \cite{rebuffi2017icarl} & DGR \cite{shin2017continual} & A-GEM \cite{chaudhry2018efficient} & MAS \cite{aljundi2018memory} & EBLL \cite{rannen2017encoder} &
HAT \cite{serra2018overcoming} & PNN \cite{rusu2016progressive} \\

ER \cite{rolnick2019experience} & CCLUGM \cite{lavda2018continual} & LL-GAN \cite{zhai2019lifelong} & IMM \cite{lee2017overcoming} & LFL \cite{jung2016less} &
Piggyback \cite{mallya2018piggyback} & RCL \cite{xu2018reinforced} \\

CoPE \cite{de2021continual} & ESGR \cite{he2018exemplar} & TAIL \cite{liu2023tail} & SI \cite{zenke2017continual} &
IADM \cite{yang2019adaptive} & DeeSIL \cite{belouadah2018deesil} & DEN \cite{yoon2017lifelong} \\ 

LL-PDR \cite{hou2018lifelong} & Hyper-LL GAN\cite{zhai2021hyper} & UPGD \cite{elsayed2024addressing} & Riemannian Walk \cite{chaudhry2018riemannian} & PAR \cite{wang2023task} & RPS-Net \cite{rajasegaran2019random} & ACL \cite{ebrahimi2020adversarial}\\
\midrule
\end{tabular*}
\end{table}

In \emph{replay methods}, samples are stored as exemplars of previous tasks or they are generated pseudo-samples through a generative model. These set of exemplars are replayed when new task is confronted with the model to alleviate forgetting of previous tasks. 
The work of \cite{chaudhry2019continual, isele2018selective, rebuffi2017icarl, rolnick2019experience, de2021continual, hou2018lifelong} these research paper shows the \emph{replay based rehearsal methods} in which the model is retrained explicitly using the exemplar set of previous tasks along with newly arrived task. Rebuffi et.al. \cite{rebuffi2017icarl} have worked in \emph{iCaRL: Incremental Classifier and Representation Learning} in which they stored the subset of each task in form of exemplar set for each class confronted previously with model and these exemplar set are used as joint training while learning the new task by the model. In \cite{rolnick2019experience}, Rolnick et. al. propose reservoir sampling as a method for data incremental learning, with the goal of keeping the number of stored samples within a predetermined budget while assuming the presence of an i.i.d. data stream.
In \cite{de2021continual} Continual Prototype Evolution (CoPE) method integrates an effective sampling strategy that is based on reservoirs with the nearest-mean classifier approach. In \cite{chaudhry2019continual} Chaudhry et. al. conduct an empirical investigation into the efficiency of a very short episodic memory in the context of a continuous learning environment in which each training example is only encountered once. Isele et. al. \cite{isele2018selective} propose an experience replay process that augments the conventional FIFO buffer and selectively stores experiences in a long-term memory in order to reduce the likelihood of people forgetting things. 
Distillation and Retrospection are two approaches used by Hou et. al. \cite{hou2018lifelong} to achieve a better balance between preservation and adaptability in lifelong learning.

 \emph{Pseudo-rehearsal}, a strategy that gives the benefits of rehearsal without actually requiring access to the previously learned material. Pseudo-rehearsal described in \cite{ramapuram2020lifelong,atkinson1802pseudo, shin2017continual, lavda2018continual, he2018exemplar, zhai2021hyper} is an alternate method that was utilised in early research using shallow neural networks when there were no previous examples available. Atkinson et. al.has described that random input cannot span the input space when dealing with deep neural networks and big input vectors (such high-resolution photos) \cite{atkinson1802pseudo}. Recently, generative models have demonstrated the capacity to produce high-quality pictures \cite{goodfellow2014generative, cong2020gan, zhai2019lifelong, zhai2020piggyback, zhang2022continual}, opening the door to modelling the data generation distribution and retraining on generated instances. However, this does add an additional layer of complexity to the process of continuously training generative models \cite{shin2017continual}. 

Rehearsal might overfit the subset of stored samples and seems to be limited by joint training, but \emph{constrained optimization} is an alternative that gives backward/forward transfer more space to maneuver. According to the principle that was presented presented by Lopez et. al in GEM \cite{lopez2017gradient} for the task incremental setting, the most important thing to do is to ensure that only new task updates do not interact with older tasks. This is accomplished by using the first order Taylor series approximation to project the predicted gradient direction onto the feasible region that was delineated by the gradients of the prior job. A-GEM \cite{chaudhry2018efficient} simplifies the issue such that it only has to project in one direction using estimates derived from samples chosen at random from the data buffer of a prior job. Aljundi et al. \cite{aljundi2019online} expand this technique to a pure online continuous learning context with no task limits, proposing to choose sample subsets that best match the global optimum of historical data. A framework for pretrained decision-making models that is efficient at adapting is presented in TAIL \cite{liu2023tail}. The author showed that parameter-efficient fine-tuning (PEFT) strategies in TAIL, particularly Low-Rank Adaptation (LoRA), can improve adaptation efficiency, reduce catastrophic forgetting, and guarantee robust performance on many tasks by conducting a thorough examination of these techniques. A new method for continuous representation learning is utility-based perturbed gradient descent, or UPGD \cite{elsayed2024addressing}. UPGD is a hybrid of gradient updates and perturbations; it protects more useful units from forgetting by applying smaller alterations to them and rejuvenates their plasticity by applying bigger modifications to less useful units.

\emph{Regularization approaches} prevent keeping raw inputs while still protecting privacy and reducing memory needs. Instead, an additional regularisation term is incorporated in the loss function to consolidate prior knowledge while learning on fresh data. These approaches are further subdivided into \enquote{prior-focused} and \enquote{data-focused} methods.

\enquote{Prior-focused} approaches \cite{kirkpatrick2017overcoming, liu2018rotate, aljundi2018memory, lee2017overcoming, zenke2017continual, chaudhry2018riemannian} estimate a distribution across the model parameters and then utilise that distribution as a prior when learning from new task. This helps to prevent forgetting. In most cases, the relevance of every neural network parameter is calculated, and it is assumed that the parameters are independent so that the network may be implemented. Changes to crucial parameters are compensated while training for later tasks. In this, first approach the \enquote*{Elastic Weight Consolidation} (EWC) method is able to recall previous tasks because it selectively slows down learning on the weights that are significant for those tasks \cite{kirkpatrick2017overcoming}. Lee et. al. has proposed \enquote*{Incremental Moment Matching} (IMM) step by step matches the moment of the neural network likelihood function, which is trained on the first and second tasks, one after the other. To make the space for searching the IMM procedure is supported by various techniques for transfer learning include weight transfer, $L2-norm$ of the old and the new parameter, and a dropout version that uses the old parameter \cite{lee2017overcoming}. Nguyen et.al. prposed Variational Continual Learning (VCL) a variational framework \cite{nguyen2017variational}, which is supported by Bayesian-based works as introduced by Ahn et. al.\cite{ahn2019uncertainty} and Zeno et. al. \cite{zeno2018task}. Estimates of significance weights have been made by Zenke et. al. online during task training \cite{zenke2017continual}. Unsupervised significance estimate was proposed by Aljundi et al. \cite{aljundi2018memory}, which would allow for additional flexibility and online user adaption as described by Lange et. al. in \cite{lange2020unsupervised}. This work was extended by Aljundi et. al. \cite{aljundi2019task} to task-free circumstances.

\emph{Data-focused} methods \cite{li2017learning, zhang2020class, rannen2017encoder, jung2016less, yang2019adaptive} start with transferring knowledge from a previous model to the model that is being trained on the new data. Silver et. al. proposed a task rehearsal method (TRM) leverages previously acquired tasks as a source of inductive bias for lifelong learning. This inductive bias helps TRM create accurate hypotheses for new problems with few training examples. TRM has a knowledge retention phase where the neural network representation of a well learned task is saved, and a knowledge recall and learning phase where virtual examples of stored tasks are created \cite{silver2002task}. \enquote*{Learning without forgetting} (LwF) method reintroduced by Li et. al. \cite{li2017learning} to decrease forgetting and transfer knowledge by using prior model output as fuzzy labels for earlier tasks. Jung et. al. \cite{jung2016less} observed that their technique makes no use of data from the source domain is quite good at forgetting less of the knowledge in the source domain, and they demonstrate its usefulness with multiple trials. Furthermore, they have discovered that forgetting occurs between mini-batches while executing broad training procedures with stochastic gradient descent techniques, and problem is that one of the variables that affects the network's generalization capability. This method has been found to be susceptible to domain shift between tasks \cite{aljundi2017expert}. Rannen et. al. \cite{rannen2017encoder} made an effort to circumvent the problem that was presented before, assist the gradual integration of shallow autoencoders to restrict task characteristics in the learnt low-dimensional space that corresponds to those features. Adaptively choosing a suitable method from parameter allocation and regularization, PAR \cite{wang2023task} takes into account the learning difficulty of each challenge. When a model has learned similar tasks, it finds them easy, and vice versa. To assess the degree of similarity between tasks utilizing just tasks-specific characteristics, the author suggests a divergence estimation technique based on the Nearest-Prototype distance.

\emph{Architectural based approach} assigns unique model parameters to each job, hence decreasing the probability of forgetting. These approaches are further sub categorised into \enquote{Fixed Network}, and \enquote{Dynamic Network} as described in Table \ref{tab:41}. When there are no size limits on an architecture, one can add new neurons for new tasks while keeping the parameters of the old tasks frozen, or for each task assign a copy of network architecture model \cite{aljundi2017expert}.
Rusu et. al. has described \enquote*{Progressive networks} that include desirable features effectively into the neural network model architecture. Catastrophic forgetting is avoided by implementing a new neural network, known as a column, for each task that is being completed, and transfer of knowledge is made possible through the use of lateral connections to characteristics of previously learned columns \cite{rusu2016progressive}. Xu and Zhu have used \enquote*{Reinforced Continual Learning} algorithms to find an optimal neural network model for each upcoming task \cite{xu2018reinforced}. In most cases, these jobs demand for the implementation of a task oracle, which is responsible for activating related masks or task branches during prediction. As a result, they are confined to a system with many heads since they are unable to successfully manage a head that is shared across jobs. To avoid this problem Aljundi et. al. have used \enquote*{Expert Gate} mechanism. They learn a specific model called an \enquote{expert} for each task by transferring knowledge from earlier tasks; more specifically, they build on the most relevant of the earlier tasks. Simultaneously, they developed a gating function that accurately represents the features of each activity. This gate sends the test data to the related expert, which improves performance on all tasks that have been learned so far \cite{aljundi2017expert}.

\emph{Fixed Network} methods such as PathNet \cite{fernando2017pathnet}, PackNet \cite{mallya2018packnet}, HAT \cite{serra2018overcoming}, Piggyback \cite{mallya2018piggyback}, DeeSIL \cite{belouadah2018deesil}, and RPS-Net \cite{rajasegaran2019random} uses regularization to ensure that any variation from an initial model is kept to a minimum when retraining models.

\emph{Dynamic Network} methods such as DAN \cite{rosenfeld2018incremental}, Expert Gate \cite{aljundi2017expert}, PNN \cite{rusu2016progressive}, RCL \cite{lange2020unsupervised}, DEN \cite{yoon2017lifelong}, and ACL \cite{ebrahimi2020adversarial} are using the network expansion.

\section{Mathematical overview of methods}
\label{sec5}
The comparison for above discussed methods as shown in table \ref{tab:41} is as following in subsections of this paper. Replay methods are compared in subsection \ref{sec5.1}, similarly Regularization methods and Architectural methods are compared in subsection \ref{sec5.2} and subsection \ref{sec5.3} respectively. 

\subsection{Replay Methods}
\label{sec5.1}
Data is either kept in its original form (raw format) or pseudo-samples are created using a generative model in this field of study. While learning a new task, these examples from prior tasks are repeated to prevent forgetting. They are either utilised again as inputs to the model for test runs, or else they are employed as constraints in the optimization of the new task loss to avoid interference from the old task. Description of replay methods is as follows:

\textbf{Tiny Episodic Memories (TEM)} \cite{chaudhry2019continual} remembers a few instances from earlier tasks and then uses these instances to training for future tasks. It uses the learning protocol used by Chaudhry et. al. \cite{chaudhry2019continual}. Following metrics are used:

\textbf{Average Accuracy} Let the performance of the proposed model on the test set of task $j$ after the model has been trained on previous task $i$ be denoted by the variable $a_{i,j}$. Then average accuracy for task $t$ is calculated as in Eq. \ref{eq51}:
\begin{equation}
    \centering
    A_{t}= \frac{1}{t}\sum_{j=1}^{t}a_{t,j}
    \label{eq51}
\end{equation}
\textbf{Forgetting} Let $f_j^i$ represent the amount of information that is forgotten on task $j$ after the model has been trained on task $i$ which may be estimated as described in Eq. \ref{eq52}:
\begin{equation}
    \centering
    f_{j}^{i}= \underset{n\in \left\{1,...,i-1 \right\}}{max} a_{n,j}-a_{i,j}
    \label{eq52}
\end{equation}
Average forgetting at task $t$ is computed as shown in Eq. \ref{eq53}: 
\begin{equation}
    \centering
    F_{t}= \frac{1}{t-1}\sum_{j=1}^{t-1}f_{j}^{t}
    \label{eq53}
\end{equation}

In \textbf{Selective Experience Replay (SER)} \cite{isele2018selective} Isele and Cosgun suggests a replay procedure that extends the functionality of the standard FIFO buffer and selectively preserves knowledge in a long-term memory. 

\textbf{Incremental Classifier and Representation Learning (iCaRL)} proposed by Rebuffi et.al. \cite{rebuffi2017icarl} keeps samples (exemplars) closest to each feature mean-of-exemplars to each class is given by Eq. \ref{eq57}:
\begin{equation}
    \centering
    \mu_y\leftarrow \frac{1}{\left| P_y\right|} \sum_{p \in P_y}\varphi \left ( p \right )
    \label{eq57}
\end{equation}
where \(\varphi : \mathcal{X} \rightarrow \mathbb{R}^d \) is feature map and $P$ is class exemplar set.
During training, projected loss on new classes and distillation loss is given by Eq. \ref{eq58} between prior and current model predictions:
\begin{equation}
\tiny
    \begin{split}
        \mathcal{L} (\Theta)=-\sum_{(x_i,y_i)\in \mathcal{D}}  \sum_{y=s}^{t}\delta_{y=y_i}log \ g_y(x_i)+\delta_{y\neq y_i}log \ (1- g_y(x_i)) \\ - \sum_{(x_i,y_i)\in \mathcal{D}} \sum_{y=s}^{s-1}q_i^y \ log \ g_y(x_i)+(1-q_i^y)log \ (1- g_y(x_i))
    \end{split}
    \label{eq58}
\end{equation}
where \( \mathcal{D}\leftarrow \bigcup_{y=s,...,t}{(x,y):x\in X^y} \ \cup \bigcup_{y=1,...,s-1}{(x,y):x\in P^y} \) is combined training set with \( \mathcal{P}=(P_1,...,P_{s-1}) \) exemplar set, $\Theta$ is network parameter, $x$ is data points, $y$ is correspondence label, and $g_y(x)$ is network output as define in Eq. \ref{eq59}:
\begin{equation}
    \centering
    g_{y}(x)=\frac{1}{1+exp(-a_y(x))}
    \label{eq59}
\end{equation}
with $a_y(x)=w_y^T \varphi(x) $, where $w$ is weights to corresponding tasks.

\textbf{Experience Replay (ER)} proposed by Rolnick et. al. \cite{rolnick2019experience} where they have used \emph{CLEAR} method that makes use of on-policy learning in addition to off-policy learning and behavioural cloning from replay in order to improve stability (old knowledge preservation) while maintaining plasticity (new knowledge acquisition). They have used \emph{V-Trace} off-policy \cite{espeholt2018impala} learning algorithm.


In \textbf{Hyper-LifelongGAN} \cite{zhai2021hyper} Zhai et. al. proposed task-specific filter generators create dynamic base filters that are then combined into a weight matrix that is shared among all tasks in sequence. Let us we assume that $M_t$ model consist of discriminator $D_t$, and generator $G_t$ with $L$ layers filter then  $\left \{ F_{t}^{t} \in \mathbb{R}^{s^{l}_{w} \times s^{l}_{h} \times c_{in}^{l} \times c_{out}^{l}} \right \} \underset {l=1}{L}$ where layers index is $l$, $s^l_w$ kernel width, $s^l_h$ is kernel height with input channels $c_{in}^l$, and output channels $c_{out}^l$. Now $F_t^l$ can be factorized into weight matrix $\mathcal{W}_t^l \in \mathbb{R}^{K \times (c_{in}^l \times c_{out}^l)}$, and base filter $\mathcal{B}_t^l \in \mathbb{R}^K \times (s_w^l \times s_h^l)$. This can be combined into Eq. \ref{eq514}:
\begin{equation}
    \centering
    F_t=\mathcal{R}(\mathcal{B}_t \ast \mathcal{W}_t)
    \label{eq513.1}
\end{equation}
where reshaping operation is denoted by $\mathcal{R}$. Now, loss of network is given at $t^{th}$ time as shown in Eq. \ref{eq514}:
\begin{equation}
    \centering
    \mathcal{L}_{total}^t \rightarrow \mathcal{L}_{task}^t + \beta \mathcal{L}_{distill}^t,
    \label{eq514}
\end{equation}
where $\beta$ constant denotes loss weight for distillation knowledge. Knowledge distillation \ref{eq515} and loss minimization of current task is calculated as \ref{eq516}:
\begin{equation}
    \centering
         \mathcal{L}_{distill}^t= \sum_{i=1}^{t-1} \left\| \widetilde{B}^{i}_{t-1}-\widetilde{B}^i_t \right\|
    \label{eq515}
\end{equation}
\begin{equation}
    \centering
     \mathcal{L}_{task}^t = \mathcal{L}_{task} \left ( A_t, B_t^t, \widetilde{B}^t_t \right )
    \label{eq516}
\end{equation}
with $\widetilde{B}^t_t=G_t^t(A_t,z)$ where $\widetilde{B}$ represent the sub-generators of model $M$.

\textbf{Gradient Episodic Memory (GEM)} \cite{lopez2017gradient} proposed by Lopez et. al. uses a set of metrics i.e. average accuracy \ref{eq517}, backward transfer\ref{eq518} and forward transfer\ref{eq519} to alleviate the catastrophic forgetting in continual learning task.
\begin{equation}
    \centering
     Accuracy = \frac{1}{T}\sum_{i=1}^{T}R_{T,i}
    \label{eq517}
\end{equation}
\begin{equation}
    \centering
     Backward \ Transfer = \frac{1}{T-1}\sum_{i=1}^{T-1}R_{T,i}-R_{i,i}
    \label{eq518}
\end{equation}
\begin{equation}
    \centering
     Forward \ Transfer = \frac{1}{T-1}\sum_{i=2}^{T}R_{i-1,i}-\overline{b}_i
    \label{eq519}
\end{equation}

where the test classification accuracy of the model on task $t_j$ after witnessing the final sample from task $t_i$ is defined as $R_{i,j}$. Each task's test accuracy can be represented by its own random-initialized vector $\overline{b}$. Now, if memory size is $M$ and known tasks are $T$, then each task occupies a memory of $m=M \setminus T$. In order to describe the loss function in the memory of the $n^th$ task, we take into account variables $f_\theta$ parameterized by $\theta \in \mathbb{R}^p$ as shown in Eq. \ref{eq520}:
\begin{equation}
    \centering
     l\left ( f_\theta, \mathcal{M}_n \right )= \frac{1}{\mathcal{M}_n}\sum_{(x_i,n,y_i)\in \mathcal{M}_n}l\left ( f_\theta(x_i,n),y_i \right )
    \label{eq520}
\end{equation}

\subsection{Regularization Methods}
\label{sec5.2}
The regularisation strength is $\mathcal{H}$ in our hyperparameter architecture since it is correlated with the degree of knowledge retention. $\mathcal{L}1$ (also called \emph{Lasso regression}), $\mathcal{L}2$ (also known as \emph{Ride Regression}), and \emph{dropout} are three effective regularisation methods used in different methodology by authors. This amount of work avoids maintaining raw inputs, and reduces memory use. Instead, an additional regularisation term is added to the loss function, which helps fresh data-driven learning to build on prior learning. Regularization methods are described as below:

\textbf{Elastic Weight Consolidation (EWC)} \cite{kirkpatrick2017overcoming}  incorporates parameter uncertainty for network parameters into the Bayesian framework. The Laplace approximation posterior is approximated as a Gaussian distribution with mean $\theta^*_{T_1}$ and diagonal precision of Fisher information Matrix (F). It is possible that there is a solution for task $T_2$, $\theta^*_{T_2}$, that is similar to the one that was previously discovered for task $T_1$, $\theta^*_{T_1}$ due to over-parameterization. EWC consequently safeguards task $T_1$ performance while learning task $T_2$ by limiting the parameters to remain in a task $T_1$ zone of low error centred on $\theta^*_{T_1}$. Given this approximation, EWC minimises $\mathcal{L}$ as in Eq.\ref{eq521}:
\begin{equation}
    \centering
     \mathcal{L}(\theta)= \mathcal{L}_{T_2}+\sum_{i}\frac{\lambda}{2}F_i\left ( \theta_i-\theta^{*}_{T_1,i} \right )^2
    \label{eq521}
\end{equation}

\textbf{Incremental Moment Matching (IMM)} \cite{lee2017overcoming} progressively matches the instant of the neural network's posterior distribution which is trained on the initial and the subsequent task, respectively. To create the search area of posterior parameter smooth, the IMM method is augmented by a dropout variation with the previous parameter and transfer learning strategies such as weight transfer, $\mathcal{L}2-norm$. While both IMM and EWC \cite{kirkpatrick2017overcoming} use model merging to arrive at projections of Gaussian posteriors for task parameters, IMM is fundamentally different due to the nature of its use of model merging. Because the merging technique anticipates using a single merged model in practice, it mandates storing models while they are being trained for each individual task. Lee et. al., have used two type of methods for merging the models: mean based IMM and mode based IMM. The weighted sum method is used as shown in Eq. \ref{eq522} to the weights $\theta_k$ of networks trained for a particular task $t$ with mixing ratio $\alpha_k^t$:
\begin{equation}
    \centering
     \theta_k^{1:T}=\sum_{t}^{T}\alpha_k^t\theta_k^t
    \label{eq522}
\end{equation}
Similarly, mode based IMM methods aims to merge the model for Gaussian mixture as shown in Eq. \ref{eq523}:
\begin{equation}
    \centering
     \theta_k^{1:T}=\frac{1}{\Omega_k^{1:T}}\sum_{t}^{T}\alpha_k^t\Omega_k^t\theta_k^t
    \label{eq523}
\end{equation}
It is worth keeping in mind that a neural network performance is unaffected by randomly selecting its nodes and weights. Due to substantial cost barriers between the parameters of the two neural networks, averaging the parameters of two neural networks started separately may result in weak performance as discussed by Goodfelow et.al. \cite{goodfellow2014qualitatively}. To overcome the above problem, Lee et. al.\cite{lee2017overcoming} have used three transfer techniques that are: \textit{Weight-transfer, Drop-transfer, and $\mathcal{L}2$-transfer}.

\textbf{Riemannian Walk (RWalk)} \cite{chaudhry2018riemannian}  measures forgetting and intransigence, which help us to better understand, analyse, and shed light on the functioning of incremental learning algorithms. With KL-divergence based approach, Chaudhary et.al introduce RWalk\cite{chaudhry2018riemannian}, an extension of Path Integral \cite{zenke2017continual}, and EWC++ \cite{kirkpatrick2017overcoming}. Rwalk consist of three main components:(1) Regularization over Conditional probabilistic likelihood based on KL-divergence, (2) Riemannian manifold (parameter importance based on sensitivity loss), and (3) exemplars from previous tasks. The objective function for RWalk is as given by Eq. \ref{eq524}:
\begin{equation}
    \centering
    \tiny
     \bar{L}^k(\theta)={L}^k(\theta)+\lambda\sum_{i=1}^{P}\left ( F_{\theta_{i}^{k-1}}+s_{t_0}^{t_{k-1}}(\theta_{i}) \right )\left ( \theta_{i}-\theta_{i}^{k-1} \right )^2
    \label{eq524}
\end{equation}
where score from $t_0$ to $t_{k-1}$ task is accumulated by $s_{t_0}^{t_{k-1}}(\theta_i)$ and $F_{\theta_{i}^{k-1}}$ denotes the Fisher matrix for $k-1$ corresponding tasks .

\textbf{Memory Aware Synapses (MAS)} \cite{aljundi2018memory} builds up a significance score for every network parameter by considering how much variation in that parameter affects the anticipated output function in unsupervised and online learning. We have a regularizer that penalises changes to parameters that are essential for earlier tasks when a new task $T_n$ has to be learnt, in addition to the new task loss $L_n(\theta)$ as shown in Eq. \ref{eq525}:
\begin{equation}
    \centering
     L(\theta)=L_n(\theta)+\lambda\sum_{i,j}\Omega_{ij}\left ( \theta_{ij}-\theta_{ij}^* \right )^2
    \label{eq525}
\end{equation}
where $\theta_{ij}^*$ is regularization term for previous sequenced task up to $T_{n-1}$ and hyperparameter is denoted by $\lambda$.

\textbf{Synaptic Intelligence (SI)} \cite{zenke2017continual} specifically want to provide particular synapse a local metric of importance in resolving problems that the network has previously been trained on tasks. Zenke et. al., have built an algorithms that monitor an \emph{important measure} $\omega_k^{\mu}$ that represents prior credit for task $\mu$ improving task objective $L_{\mu}$(as shown in Eq. \ref{eq526}) to individual synapses $\theta_k$ and regularization strength is calculated as Eq. \ref{eq527}:
\begin{equation}
    \centering
     \tilde{L}_{\mu}=L_{\mu}+\underbrace{c\sum_{k}\Omega_k^{\mu}\left ( \tilde{\theta}_k-\theta_k \right )^2}_{surrogate\ loss}
    \label{eq526}
\end{equation}

\begin{equation}
    \centering
     \Omega_k^{\mu}=\sum_{\nu<\mu}\frac{\omega_{k}^{\nu}}{(\Delta_{\mu}^{\nu})^2+\xi}
    \label{eq527}
\end{equation}

\textbf{Learning without Forgetting (LwF)} \cite{li2017learning} By using knowledge distillation \cite{hinton2015distilling}, LwF keeps track of previous work. The network outputs for the newly arrived task data are recorded prior to training the new task, and they are then utilised to deduce previous task knowledge during training. The effectiveness of this strategy, however, largely relies on correlation of the new task data and to earlier tasks. As additional different tasks are introduced, distribution adjustments with regard to the previously acquired activities may cause a progressive erroneous build-up to the earlier tasks \cite{aljundi2017expert}. As shown by Rebuffi et. al., \cite{rebuffi2017icarl} the same error build-up holds true for class-incremental setups. The extra work involved in transmitting all new task data and storing the results is another downside. Although LwF was developed with classification in mind, it has found use in other contexts, such as object identification \cite{shmelkov2017incremental}. Li Zhizhong and Hoiem Derek has calculated shared parameter $\theta_s^{\ast}$, task specific parameters for each old task $\theta_o^{\ast}$, and randomly initialization of new parameters to new task $\theta_n^{\ast}$ as shown in Eq. \ref{eq528}: 
\begin{dmath}
    \centering
     \theta_s^*, \theta_o^*, \theta_n^* \leftarrow  \underset{\theta_s^*, \theta_o^*, \theta_n^*}{argmin}\left ( \mathcal{L}_{old}(Y_o,\hat{Y}_o)+\mathcal{L}_{new}(Y_n,\hat{Y}_n)+\mathcal{R}(\hat{\theta}_s, \hat{\theta}_o, \hat{\theta}_n) \right )
    \label{eq528}
\end{dmath}
The multinomial logistic loss \cite{krizhevsky2017imagenet,simonyan2014very} in multiclass classification is defined as $\mathcal{L}_{new}(y_n,\hat{y}_n)=-y_n\cdot log\ \hat{y}_n$ and knowledge distillation loss used by Hinton et. al., \cite{hinton2015distilling} is calculated as  $\mathcal{L}_{old}(y_o,\hat{y}_o)= -\sum_{i=1}^{l}{y}^{'(i)}_o log \ \hat{y}^{'(i)}_o$
 where number being labelled by $l$ and $y^{'(i)}_o$ and $\hat{y}^{'(i)}_o$ are refined versions of previously-recorded and currently-applicable probability $y^{(i)}_o$ and $\hat{y}^{(i)}_o$.

\textbf{Encoder Based Lifelong Learning (EBLL)} \cite{rannen2017encoder} solves each task by optimising an autoencoder from beginning to finish, which maps features around on a lower-dimensional space. During training, a regularisation term is included to reduce the likelihood that future feature projections would deviate from the previously learned, task-optimal values. The size of the autoencoder is just a minor part of the network, but the amount of memory needed rises in proportion to the number of tasks. The primary computational burden is in autoencoder training and accumulating feature projections for every optimization phase. 

In \textbf{Less-forgetting Learning (LFL)} \cite{jung2016less}, the top layer is considered as a linear classifier, whereas the lower layer is regarded as feature extractor. As a result, the softmax function weights serve as a decision boundary for labeling the features. Jung et. al., have used two properties to handling catastrophic forgetting: (1) There is no change in decision boundary, and (2) Target network features should be near to source network characteristics. They train the network to minimise total loss as in Eq. \ref{eq530}:
\begin{equation}
    \centering
     \mathcal{L}_t(x;\theta^{(s)},\theta^{(t)})=\lambda_c\mathcal{L}_c(x;\theta^{(s)})+\lambda_e\mathcal{L}_e(x;\theta^{(s)})
    \label{eq530}
\end{equation}
where tuning parameters are denoted by $\lambda_e$, and $\lambda_c$. Further, $\mathcal{L}_t$ is total loss, $\mathcal{L}_c$ is cross entropy loss, and Euclidean loss function is denoted by $\mathcal{L}_e$.

\textbf{Adaptive Deep Models for Incremental Learning (IADM)} \cite{yang2019adaptive} deals with the challenge of streaming data that are \emph{Capacity Scalability} \cite{ronneberger2015u,yang2018complex,feng2018multi} and \emph{Capacity Sustainability} \cite{chaudhry2018riemannian,kirkpatrick2017overcoming,lee2017overcoming} (concept drift). IADM can adapt from a shallow network (rapid convergence) to a deep model (high capacity), improving overall prediction performance. The adaptive regularisation is expressed as in Eq. \ref{eq531} and whole loss function is calculated as in Eq. \ref{eq532}:
\begin{equation}
    \centering
     R=\frac{1}{T}\sum_{t=2}^{T}\sum_{i}\alpha_{t-1}\odot F_{{\theta_{t-1}}_i}({\theta_{t}}_i-{\theta^\ast_{t-1}}_i)^2
    \label{eq531}
\end{equation}
In the fisher information matrix, $\odot$ is multiplying the $\alpha_l$ by the corresponding layer parameters.
\begin{equation}
    \centering
    \tiny
     \mathcal{L}=l_t\left ( \sum_{l=1}^{L}\alpha_l f_l(x),y \right ) + \frac{\lambda}{T} \sum_{i}\alpha_{t-1}\odot F_{{\theta_{t-1}}}({\theta_{t}}_i-{\theta^\ast_{t-1}})^2
    \label{eq532}
\end{equation}
Here, gradients tends to zero when a local minimum is at $\theta_{t-1}$ making fisher matrix $F_{\theta_{t-1}}$ very small.

\subsection{Architectural Methods}
\label{sec5.3}
By locking in the subsets of parameters used in earlier tasks, this family of approaches may provide the maximum possible stability for individual tasks. While this prevents stability degradation, it also allows the available resources to be utilised as efficiently as possible, preventing capacity saturation and maintaining steady learning for subsequent tasks. Description of architectural methods is given below:

\textbf{PathNet} \cite{fernando2017pathnet} is a neural network approach that use agents implanted in the network to determine which sections of the network to reuse for new tasks. $P$ genotypes (paths) are generated at random, and each genotype is little more than a $N$ by $L$ matrix of numbers describing the active units in each layer of that pathway. A genotype is selected at random, and its route is trained for $T$ epochs. The fitness of the genotype is measured by the amount of negative classification error it experienced throughout the training period. Then, another genotype's route is trained for $T$ epochs.
Winner pathway genotype overwrites loser pathway genotype. The genotype of the successful route is then mutated by separately selecting each element with $1/[N \times L]$ probability.

\textbf{PackNet} \cite{mallya2018packnet} uses incremental pruning and model retraining, where one may \enquote{pack} several tasks into a common network with minimal drop performance and storage impact. Every convolution layer and fully connected layer has a predetermined number of eligible weights removed in each pruning phase. The lowest 50\% or 75\% of the weights within a layer are chosen for elimination after the weights are sorted by absolute magnitude. Only weights associated with the present task are pruned; weights associated with a previous task are left intact. The pruning mask protects task performance by ensuring that the task parameter subset is fixed for subsequent tasks.

\textbf{Hard Attention to the Task (HAT)} \cite{serra2018overcoming} remembers completed tasks without interfering with future learning. Using backpropagation and minibatch stochastic gradient descent, we learn a task alongside almost-binary attention vectors using gated task embeddings. Previous task attention vectors are used to construct a mask that limits how the network's weights may change in response to new input. The gating mechanism purpose is to create rigid attention masks, which may be binary in nature. These attention masks function as "inhibitory synapses," and as a result, they have the ability to activate or deactivate the output of the units in every layer. In order to do this, authors have first add a regularisation term as in Eq.\ref{eq534} to the loss function $\mathcal{L}$ as shown in Eq. \ref{eq533}, which considers the set of accumulated attention vectors up to task t-1 as:
\begin{equation}
    \centering
     \mathcal{L}'(y,\hat{y},A^t,A^{<t}) = \mathcal{L}(y,\hat{y})+\alpha R(A^t,A^{<t})
    \label{eq533}
\end{equation}
where $\alpha$ is a regularization constant, and for $t-1$ task cumulative attention vector is denoted as $A^{<t}=\{a_1^{<t},...,a_{L-1}^t\}$.
\begin{equation}
    \centering
     R(A^t,A^{<t})= \frac{\sum_{l=1}^{L-1}\sum_{i=1}^{N_t}a^t_{l,i}(1-a^{<t}_{l,i})}{\sum_{l=1}^{L-1}\sum_{i=1}^{N_t}(1-a^{<t}_{l,i})}
    \label{eq534}
\end{equation}
where $a^t_{l,i}$ is hard attention values for $t$ task, number of unit in layer $l$ is denoted by $N_l$.

\textbf{Piggyback} \cite{mallya2018piggyback} approach involves learning to selectively mask a base network fixed weights in order to enhance performance on a new task. To do this, Mallya et. al., keep a set of legitimate weights that are subjected to a deterministic adaptive threshold function to produce binary masks, which are then applied to the weights that already exist. We want to learn binary masks suitable for the task at hand by backpropagating the real-valued weights. We may reuse the same underlying base network for several tasks with little effort by learning unique binary-valued masks for each task that are applied element-wise to network parameters. Even if we do not change the network weights, masking may be used to create a wide variety of different filters. By applying a strict binary threshold function, as shown in Eq. \ref{eq535.1}, to the mask weight matrices $m^r$, we get threshold mask matrices with $\mathcal{T}$ selected threshold:
\begin{equation}
    \centering
     m_{ji}=\left\{\begin{matrix}
1, \  \ if \  m^r_{ji}> \mathcal{T} & \\ 
0, \ \ \ otherwise & 
\end{matrix}\right.
    \label{eq535.1}
\end{equation}
For these fully-connected mask weights $m$ at the threshold, we get the backpropagation equation as shown in Eq. \ref{eq535.2}:
\begin{equation}
    \centering
     \therefore \delta m \overset{\triangle}{=} \left [ \frac{\partial E}{\partial m} \right ]_{ji}=(\delta y \cdot x^T)\odot W
    \label{eq535.2}
\end{equation}
considering the error function $E$ and its partial derivative with respect to a given variable $y$, denoted by $\delta y$.

\textbf{Deep-Shallow Incremental Learning (DeeSIL)} \cite{belouadah2018deesil} is a modification of a well-known transfer learning method that increases recognition capacity by combining a fixed deep representation that serves as a feature extractor with the learning of independent shallow classifiers. Because each new notion can be introduced in under a minute and works well with a little memory budget, this technique addresses the two problems outlined above. Furthermore, since no deep re-training is required when the model is trained incrementally, DeeSIL may include larger numbers of beginning data, resulting in more transferable characteristics.

\textbf{Random Path Selection (RPS-Net)} \cite{rajasegaran2019random} promotes parameter sharing while gradually choosing the most suitable pathways for the new tasks. Because of reusing earlier pathways allows the transmission of information in a forward direction, this method necessitates a lower level of computing overhead. A path $P_k \in \mathbb{R}^{M \times l}$ can be defined for a task $k$ for a particular \emph{RPS-Net} with $M$ modules and $L$ layers as given by \ref{eq537.1}:
\begin{equation}
    \centering
     P_k(l,m)=\left\{\begin{matrix}
1, & if \ module \ \mathcal{M}_m^l & is \ added \ to \ path\\ 
0, & otherwise & 
\end{matrix}\right.
    \label{eq537.1}
\end{equation}
A loss function that incorporates both the conventional cross-entropy loss and a to train a network in small increments using distillation loss. Rajasegaran et. al., computed the cross-entropy loss for a set of $k \in [1,K]$ tasks, where each task has $N$ classes, as follows \ref{eq537.3}:
\begin{equation}
    \centering
    \tiny
     \mathcal{L}_{cross \ entropy}=-\frac{1}{n}\sum_{i}t_i[1:k \ast N]log(softmax(q_i[1:k \ast N]))
    \label{eq537.3}
\end{equation}
Authors also applied distillation loss \ref{eq537.4} in objective function to make the network resistant to catastrophic forgetting.
\begin{dmath}
    \mathcal{L}_{distillation}=\frac{1}{n}\sum_{i}KL\left ( log\left ( \sigma \left ( \frac{q_i[1:(k-1) \ast N]}{t_e} \right ) \right ), \\ \sigma \left ( \frac{q'_i[1:(k-1) \ast N]}{t_e} \right ) \right )
    \label{eq537.4}
\end{dmath}


Now hybrid loss is calculated as $\mathcal{L}=\mathcal{L}_{cross \ entropy}+\phi(k,\gamma)\cdot \mathcal{L}_{distillation}$ where scaling factor is represented by $\gamma$, and scalar coefficient is $\phi(k,\gamma)$.

\textbf{Deep Adaptation Network (DAN)} \cite{rosenfeld2018incremental}, where the filters that are learnt are limited to linear combinations of the ones that have already been learned. When compared to traditional fine-tuning processes, DANs use fewer parameters and converge to an equivalent or superior level of performance in fewer training cycles due to their capacity to retain performance on the original domain with pinpoint accuracy. When used with normal network quantization approaches, we may further lower the original parameter cost while sacrificing little or no accuracy. It is possible to tackle a problem from numerous domains using a single network by instructing the learnt architecture to transition between distinct learned representations.The output of accumulated layer $l$ is calculated as \ref{eq538}:
\begin{equation}
    \centering
     x_{l+1}=\sum_{i=1}^{n}\alpha_i(F_l^{a_i}\ast x_l+b_l^i)
    \label{eq538}
\end{equation}
where for $i_{th}$ task adapted bias are $b_l^i$, $F_l^{a_i}$, and $\alpha \in \{ 0,1 \}^n$ is vector for $n$ task.

\textbf{Expert Gate (EG)} \cite{aljundi2017expert} guarantee the scalability of this process, past task data cannot be kept and is thus unavailable while learning a new task. In such a situation, the choice of which expert to deploy throughout the test is crucial. Aljundi et.al., offer a collection of gated autoencoders that automatically send the test sample to the appropriate expert at test time after learning a representation for the current task. At any one moment, only one expert network must be put into memory, resulting in memory efficiency. Furthermore, autoencoders automatically record the relatedness of one task to another, allowing the most relevant previous model to be picked for fine-tuning or learning-without-forgetting when training a new expert.
Each task autoencoder degree of certainty is represented as a probability $p_i$ by the softmax layer with reconstruction error $er_i$ and $t$ temperature as shown in Eq. \ref{eq539.1}, and how closely the two tasks are related to each other is computed as in Eq. \ref{eq539.2}:
\begin{equation}
    \centering
     p_i=\frac{exp(-er_i/t)}{\sum_{j}exp(-er_j/t)}
    \label{eq539.1}
\end{equation}

\begin{equation}
    \centering
     Relatedness(T_k,T_a)=1-\left ( \frac{er_a-er_k}{er_k} \right )
    \label{eq539.2}
\end{equation}

\textbf{Progressive Neural Networks (PNN)} \cite{rusu2016progressive}: One of the primary aims of machine learning is lifelong learning process, where agents not only acquire knowledge from a sequence of tasks encountered in order, but also apply that knowledge to new problems, hence increasing the rate of convergence. These requirements are built into the structure of a progressive network from the start: to avoid catastrophic forgetting, a new neural network is created for each task, and lateral connections to features from previously learned columns allow for transfer. Hidden activation $h_i^{(k)}$ is calculated by for $k$ task is as \ref{eq540}:
\begin{equation}
    \centering
     h_i^{(k)}=f\left ( W_i^{(k)}h_{i-1}^{(k)}+\sum_{j<k}U_i^{(k:j)}h_{i-1}^{(j)} \right )
    \label{eq540}
\end{equation}
where weight matrix for layer $i$ is of task $k$ is denoted by $W_i^{(K)}$, and lateral connections from previous layers is denoted  by $U_i^{(k:j)}$.

\textbf{Reinforced Continual Learning (RCL)} \cite{xu2018reinforced} uses three distinct networks: the \emph{controller network, the value network, and the task network}. A Long Short-Term Memory network (LSTM) is used as the controller to generate rules and decide how many filters or nodes should be added for each activity. To estimate the value of a state, we model it as a fully-connected network, which we refer to as the value network. Any network of relevance for accomplishing a specific goal, such as image classification or object identification, may be used as the task network.

\textbf{Dynamically Expandable Networks (DEN)} \cite{yoon2017lifelong} is able to dynamically determine its network capacity while it trains on a succession of tasks, with the goal of learning a compact knowledge sharing structure that overlaps between the tasks. DEN is successfully able to avoid semantic drift by dividing and duplicating units and timestamping. DEN is trained in an effective way online by conducting selective retraining. DEN dynamically grows network capacity upon the arrival of each task with just the required number of units. At time $t$, the objective of the lifelong learning agent is to learn the model parameter $W^t$ by finding a solution to the following Eq. \ref{eq542.1}:
\begin{equation}
    \centering
     \underset{w^t}{minimize} \ \mathcal{L}(W^t;W^{t-1},\mathcal{D}_t)+\lambda \Omega(W^t)
    \label{eq542.1}
\end{equation}
Retraining the model for each new task would be naive. Deep neural network retraining is expensive. The authors Yoon et. al., advise selectively retraining just the weights impacted by the new task. Sparsity in the weights is encouraged by training the network using $l1$-regularization; this means that each neuron is only connected to a small number of neurons in the layer as shown in Eq. \ref{eq542.2}.
\begin{equation}
    \centering
     \underset{W^{t=1}}{minimize} \ \mathcal{L}(W^{t=1}; \mathcal{D}_t)+\mu \sum_{l=1}^{L}\left \| W_l^{t=1} \right \|_1
    \label{eq542.2}
\end{equation}
If the new task is extremely related to the previous ones, or if the accumulated partial knowledge received from each task is adequate to understand the new task, selective retraining will sufficient. When the learnt features are unable to adequately reflect the new task, extra neurons must be inserted into the network to account for the characteristics required for the new task. The authors Yoon et. al., suggest employing group sparse regularisation to dynamically determine how many neurons to add at which layer for each task without retraining the network for each unit as shown in Eq. \ref{eq542.5}:
\begin{equation}
    \centering
    \tiny
     \underset{W_{l}^{\mathcal{N}}}{minimize} \ \mathcal{L}(W_{l}^{\mathcal{N}};W_{l}^{t-1},\mathcal{D}_t) + \mu\left \| W_l^{\mathcal{N}} \right \|_1 + \gamma \sum_{g}\left \| W_{l,g}^{\mathcal{N}} \right \|_2
    \label{eq542.5}
\end{equation}
Semantic drift, or catastrophic forgetting, is an issue in lifelong learning when the model increasingly adapts to later acquired tasks and forgets what it learned for previous tasks, resulting in degenerate performance. The most common and straightforward technique to avoid semantic drift is to regularise parameters using l2-regularization as shown in Eq. \ref{eq542.6}:
\begin{equation}
    \centering
     \underset{W^{t}}{minimize} \ \mathcal{L}(W^{t};\mathcal{D}_t) + \lambda \left \| W^{t}-W^{t-1} \right \|_2^2
    \label{eq542.6}
\end{equation}

\textbf{Adversarial Continual Learning (ACL)} \cite{ebrahimi2020adversarial}  proposes a hybrid continuous learning system that learns task-invariant and task-specific characteristics. The authors Ebrahimi et. al., proposed model that combines architectural evolution and experience replay to avoid task-specific forgetting.

\section{Discussion and Conclusion}
\label{sec6}
We can clearly observe that in Table \ref{tab:32}, where all three scenarios of incremental learning performed on split MNIST dataset; the performance of Class based learning scenario and Domain based learning scenarios is not promising and accuracy of learning is also very low. However, Domain based incremental learning has an edge to Class based learning scenario. If we compare both these scenario for different method we can get that in baseline(lower bound) the accuracy difference is about $\approx40\%$ and baseline(upper bound)- $\approx1\%$. For regularization approach in which EWC, Online EWC, and SI methods shows a clear difference of about $\approx42\%$ in domain based and class based. However, for Replay approach in which the difference between methods of Domain based learning and Class based learning is as follows: LwF($\approx48\%$), DGR($\approx5\%$), DGR+distill($\approx1\%$) and Replay+Exemplars for class based learning it is $\approx 93.97\%$. This observation shows that domain based and class based learning scenarios performance is not promising for incremental learning. However, if we observe the Table \ref{tab:32}, we can see that for Task based learning it has promising result (more than $\approx95\%$) in all approaches of incremental learning. So we can conclude that for split MNIST dataset, Task based learning scenario performs well in terms of accuracy. Similarly, for p-MNIST dataset, all approaches performs well for Task based learning and Domain based learning except LwF method(see Table \ref{tab:33}). For Class based incremental learning regularization methods did not produces good results. Further, we can see from Table \ref{tab:41} in which we have described different approaches for incremental learning. For every approach we have given abstract details of  each methods with their loss functions and accuracy. 

\section{Future work}
\label{sec7}

The development of lifelong learning systems that can communicate with other databases is another promising area of study. This capacity becomes extremely important when dealing with real-world interactive systems, such as embodied agents or dialogue systems. Because of the dynamic nature of the world, these interactive systems must be prepared to deal with novel subjects, objects, and items. A static language model trained at a single instant cannot, for instance, respond to the question "what is the current air quality in Prayagraj, India" in the absence of an external knowledge base that can supply it with the current air quality in Prayagraj, India. One approach to overcoming this obstacle is to create systems for continuous learning that focus on skill acquisition rather than knowledge acquisition. These systems could learn how to query external knowledge-bases, aggregate results from multiple searches, etc. A common method for imparting such knowledge is to compile a set of activities, either formally or informally organized into a curriculum, the completion of which compels the system to either acquire new knowledge or refine its application of existing knowledge.








\end{document}